\documentclass[sn-mathphys]{sn-jnl}
\usepackage{xcolor}

\jyear{2021}%

\theoremstyle{thmstyleone}%
\theoremstyle{thmstyletwo}%

\theoremstyle{thmstylethree}%

\begin{document}

\title[How to identify appropriate Dataset Splits]{Surgical Phase and Instrument Recognition: How to identify appropriate Dataset Splits}

\author*[1,2]{\fnm{Georgii} \sur{Kostiuchik}}\email{georgii.kostiuchik@med.uni-heidelberg.de}

\author[1,2]{\fnm{Lalith} \sur{Sharan}}

\author[3]{\fnm{Benedikt} \sur{Mayer}}

\author[4]{\fnm{Ivo} \sur{Wolf}}

\author[3]{\fnm{Bernhard} \sur{Preim}}

\author[1,2]{\fnm{Sandy} \sur{Engelhardt}}

\affil[1]{\orgdiv{Department of Cardiac Surgery}, \orgname{Heidelberg University Hospital}, \orgaddress{\city{Heidelberg}, \country{Germany}}}

\affil[2]{\orgname{DZHK (German Centre for Cardiovascular Research)}, \orgaddress{partner site Heidelberg/Mannheim}, \country{Germany}}

\affil[3]{\orgdiv{Department of Simulation and Graphics}, \orgname{University of Magdeburg}, \orgaddress{\city{Magdeburg}, \country{Germany}}}

\affil[4]{\orgdiv{Department of Computer Science}, \orgname{Mannheim University of Applied Sciences}, \orgaddress{\city{Mannheim}, \country{Germany}}}

\abstract{\textbf{Purpose:} Machine learning approaches can only be reliably evaluated if training, validation and test data splits are representative and not affected by the absence of classes that are of interest. Surgical workflow and instrument recognition are two tasks that are complicated in this manner, because of heavy data imbalances resulting from different length of phases and their potential erratic occurrences. Furthermore, the issue becomes difficult as sub-properties that help to define phases, like instrument \mbox{(co-)occurrence}, are usually not particularly considered when defining the split. We argue that such sub-properties must be equally considered. 

\textbf{Methods:} We present a publicly available data visualization tool that enables interactive exploration of dataset partitions for surgical phase and instrument recognition. The application focuses on the visualization of the occurrence of phases, phase transitions, instruments, and instrument combinations across sets. Particularly, it facilitates assessment of dataset splits for surgical workflow recognition, especially regarding identification of sub-optimal dataset splits.

\textbf{Results:} To validate the dedicated interactive visualizations, we performed analysis of the common Cholec80 dataset splits using the proposed application. We were able to uncover phase transitions, and combinations of surgical instruments that were not represented in one of the sets. Addressing these issues, we identify possible improvements of the splits using our tool. A user study with ten participants demonstrated that the participants were able to successfully solve a selection of data exploration tasks using the proposed application. 

\textbf{Conclusion:} In highly unbalanced class distributions, special care should be taken with respect to the selection of an appropriate dataset split. Our interactive data visualization tool presents a promising approach for the assessment of dataset splits for the tasks of surgical phase and instrument recognition. Evaluation results show that it can potentially enhance the development process of machine learning models. The live application is available at \url{https://cardio-ai.github.io/endovis-ml/}.
}


\maketitle

\section{Introduction}\label{introduction} 
Technologies that enable next-generation context-aware systems in the operating room are currently intensively researched in the domain of surgical workflow recognition \cite{MaierHein2017}. Recent studies that apply machine learning algorithms to this task have shown the most promising results \cite{Garrow2021}. To further support advances in this area, academic machine learning competitions are hosted regularly \cite{MaierHein2021,Wagner2023}. However, despite the progress in surgical workflow recognition, the developers of machine learning algorithms are faced with several challenges that result from the heterogeneous nature and complexity of surgical workflows, and the temporal correlation of sensor data.

Specifically, one of the major challenges of surgical workflow data lies in the unequal distribution of classes (i.\,e., surgical phases), which is commonly referred to as data imbalance in the machine learning literature. This issue is further exacerbated by the fact that some phases can occur several times during surgery while other phases may not occur at all. This results in an imbalanced representation of classes in the dataset which in turn hinders the ability of machine learning classifiers to accurately predict the underrepresented classes. To ensure that a machine learning model can properly learn to discriminate surgical phases, all dataset splits into train, validation, and test sets must follow similar distributions. Besides, the surgical phases strongly correlate with the instruments that are used during the phase. Therefore, unequal distribution of phases also affects the distribution of sub-properties in the datasets, such as surgical instruments.

In this work, we present an interactive data visualization application that facilitates the assessment of dataset splits for surgical phase and instrument recognition with regard to the aforementioned challenges. The main goal of this work is to provide a data visualization tool that can be used by machine learning practitioners as well as biomedical challenge organizers to gain insights into dataset splits of surgical workflow data.

\section{Related work}
With the advent of deep learning, the topics of automatic phase and instrument recognition have gained considerable traction. In one of the earliest studies on this topic, \mbox{Twinanda et al.} \cite{Twinanda2017} fine-tune a convolutional neural network for joint phase and instrument recognition and apply a hidden Markov model to enforce temporal dependencies of phase predictions. Another study by \mbox{Jin et al.} \cite{Jin2018} present an improvement upon the previous work by training a deep convolutional network and a recurrent neural network in an end-to-end manner. Recently, a multi-stage temporal convolutional network has been successfully applied to the task of surgical phase recognition by \mbox{Czempiel et al.} \cite{Czempiel2020}. In the latest studies, the focus has shifted towards the transformer architectures \cite{Czempiel2021,Gao2021,Jin2022,Zou2022,Pan2023}.

Data visualization techniques represent a promising approach that can facilitate the exploration of surgical workflows. Yet, only limited research on visualization techniques for the analysis of surgical workflows has been conducted so far. Previously, \mbox{Blum et al.} \cite{Blum2008} used a hidden Markov model to derive a workflow model from a set of procedures and visualize it as a graph. One of the most recent studies by \mbox{Mayer et al.} \cite{Mayer2023} presents an interactive visualization method that focuses on the analysis of temporal relationships within the surgical workflow data and provides means for comparing sets of procedures (e.g., stratified by surgeon, pathology, etc.).

\section{Visualization Framework}
The proposed visualization framework aims to facilitate interactive exploration of dataset splits for surgical workflow recognition. The scope of this work is limited to the analysis of surgical phase and instrument annotations. This data contains information that is crucial for creating representative dataset splits. With this data, information about the duration of surgeries, the occurrence of phases, as well inter- and intra-phase instrument usage can be directly inferred. Furthermore, transitions between phases, co-occurrence of surgical instruments, and idle segments of surgeries also need to be considered during the preparation of the dataset split.

The framework comprises two main views that focus on the visualization of surgical phases and instruments. Further supplementary views provide a general overview of the dataset. The colors red, green, and blue are used consistently across all views to encode attributes of the training, validation, and test set respectively. In this section, we use eight proctocolectomy surgeries from the ``Surgical Workflow Analysis in the sensorOR 2017'' challenge dataset \cite{MaierHein2021} and also select frames that are annotated with both phases and instruments. The live application can be accessed at \url{https://cardio-ai.github.io/endovis-ml/}.

\subsection{Phase view}
\begin{figure}[h]
\centering
\includegraphics[width=\textwidth]{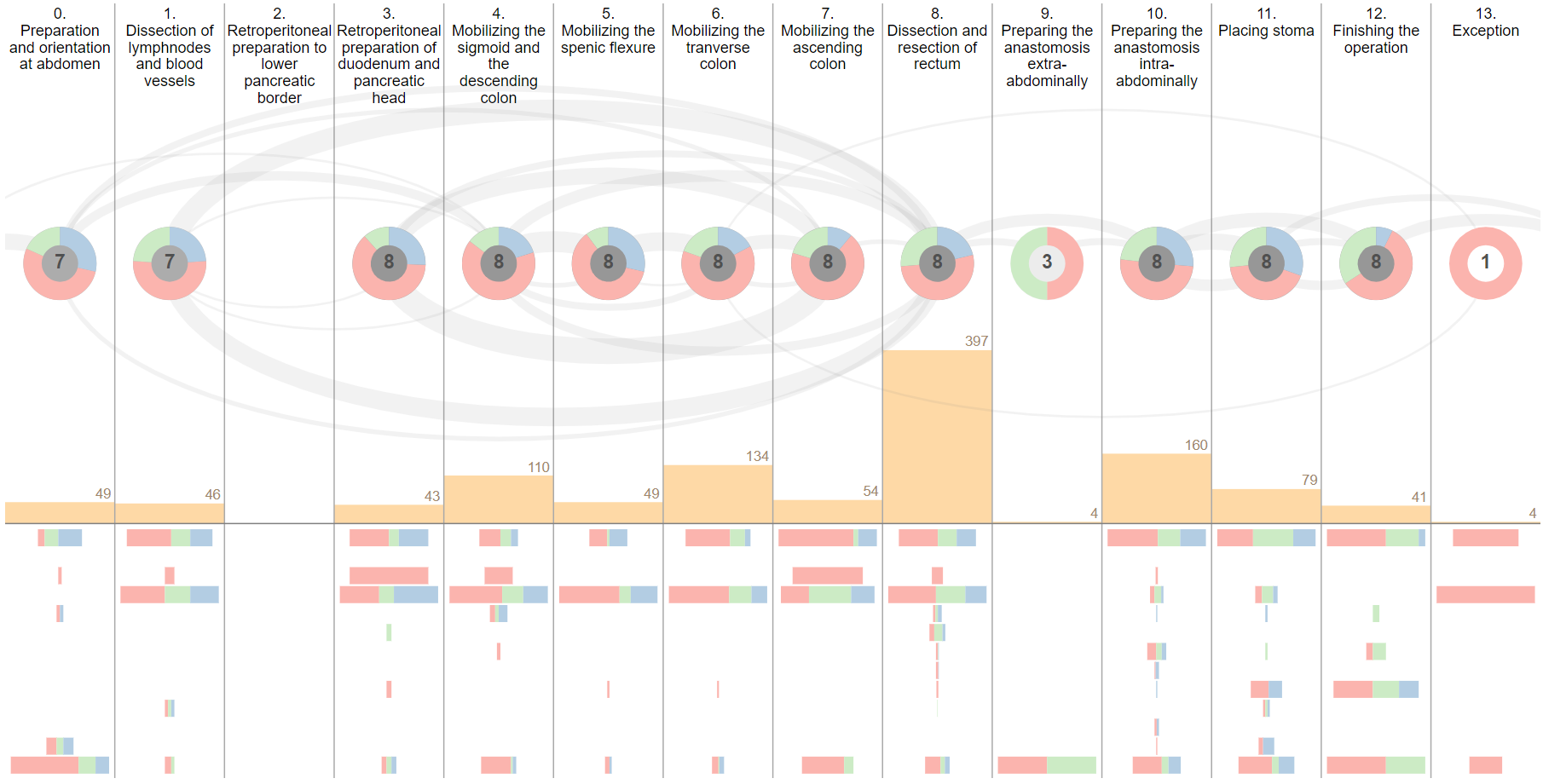}
\caption{Phase view of the proposed application with eight proctocolectomy surgeries from the ``Surgical Workflow Analysis in the sensorOR 2017'' challenge dataset.}\label{phase_view}
\end{figure}

In this view, phases are visualized as nodes along the horizontal axis, ordered according to their conceptual order from left to right (see Figure~\ref{phase_view}). Each node contains a donut chart that represents the proportion of frames that are assigned to the corresponding dataset split. Furthermore, the center of each node shows the number of surgeries in which the phase occurs. Phase transitions are visualized as arcs between individual nodes, whereas the number of times a transition between two phases happened is mapped to the width of arcs. Since transitions can occur in both directions, forward transitions are displayed in the upper half, while backward transitions are placed in the lower half of the chart. The overall distribution of frames across surgical phases is displayed as a bar chart below the phase nodes. Finally, the horizontal bars at the bottom of the view visualize the frequency of instrument occurrences during each phase and can be re-scaled in various ways depending on the analysis goal.

In order to support interactive exploration of the data, several interaction techniques are implemented in the phase view. By selecting individual phase nodes, filtering is applied across other views to display frames for the selected set of phases. Furthermore, surgeries can be filtered by the occurrence of a particular phase transition. The phase view and other views are updated accordingly to display the surgeries that contain the selected transition. Besides, the occurrence of phase transitions in the training, validation, and test sets can be displayed upon selecting the corresponding option in the phase view menu.

\subsection{Instrument view}
\begin{figure}[h]
\centering
\includegraphics[width=\textwidth]{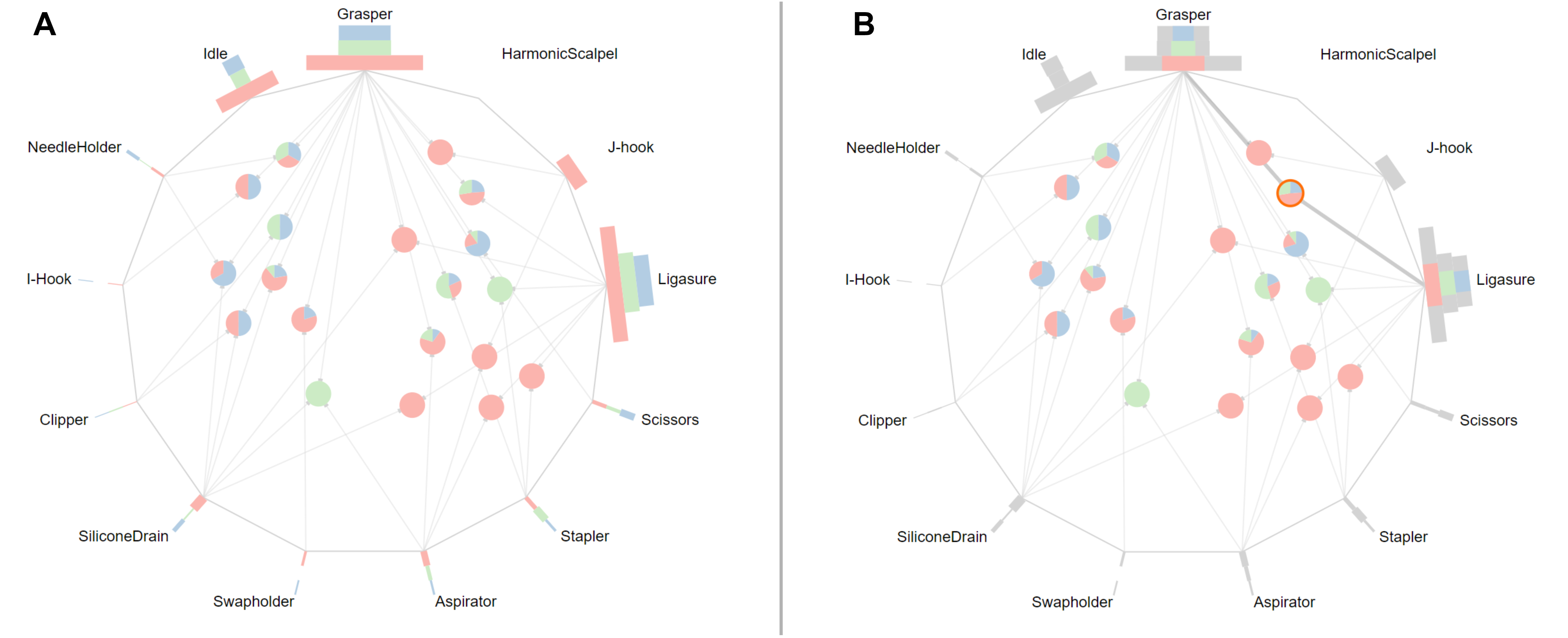}
\caption{Instrument view of the proposed application with eight proctocolectomy surgeries from the ``Surgical Workflow Analysis in the sensorOR 2017'' challenge dataset (A) and selected combination of grasper and ligasure (B).}
\label{instrument_view}
\end{figure}

The instrument view targets the visualization of instruments as well as the combinations of instruments that have been used at the same time, i.\,e. instrument co-occurrences (see Figure~\ref{instrument_view}a). This visualization approach is based on the work by \mbox{Alsallakh et al.} \cite{Alsallakh2013} which targets analysis of set memberships of data elements. The centered bar charts which are arranged radially show the total number of frames a surgical instrument was visible in each set. Additionally, a bar chart that reflects the number of frames in which no instruments are visible, so-called idle frames, is also included in this view. The combinations of instruments are displayed as nodes in the center of the instrument view. The nodes themselves are represented as pie charts, whereas each segment of the pie chart shows the prevalence of this instrument combination in the training, validation, and test set. The positioning of the nodes is determined by a force-directed layout algorithm.

To facilitate the exploration of the surgical instrument data, several interaction techniques are implemented in this view. By selecting an instrument, all instrument co-occurrence nodes that involve the selected instrument are highlighted in the instrument view. Besides, co-occurrence nodes can be selected individually which reveals the proportion of co-occurrence frames in relation to the frames of the involved instruments (see Figure~\ref{instrument_view}b). Upon filtering of instruments or instrument co-occurrences, other views of the visual framework are updated accordingly to view the selected frames.

\subsection{Supplementary views}
The main views are enhanced by two supplementary views which provide a general overview of the dataset. The first supplementary view represents a table that shows the partitioning of surgeries into the training, validation, and test sets. The individual surgeries can be interactively re-assigned to a different set via drag and drop. The second supplementary view encompasses two bar charts that display the total number of surgeries and frames for each set. Additionally, a set of bar charts displaying the number of frames for each individual surgery are arranged on the right side of the view. The average number of frames for each set are shown as dashed lines in the bar charts.

\section{Evaluation and results}\label{evaluation_results}
The proposed visualization framework is evaluated through a user study which is performed on a different dataset, namely Cholec80 \cite{Twinanda2017}. In addition to the user study, we perform analysis of the most commonly used dataset splits of the Cholec80 dataset using the proposed visualization framework.

\subsection{User study}
\begin{figure}[h]
\centering
\includegraphics[width=\textwidth]{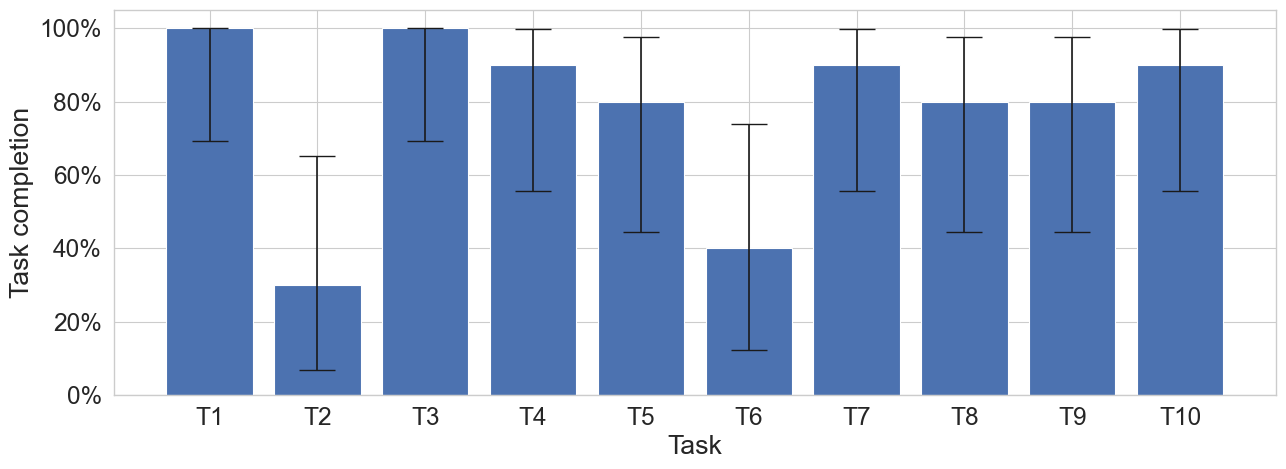}
\caption{Overall task completion percentage with the corresponding 95\%-confidence intervals.}\label{user_study_results}
\end{figure}

In total, ten participants with data science background have been recruited to participate in the evaluation study of the proposed visualization framework. After a brief introduction into the domain of surgical phase recognition, the participants were asked to solve ten tasks covering a wide range of possible exploratory analyses that can arise during the preparation of a dataset for surgical phase recognition. Descriptions of the tasks are enclosed in the supplementary information. To measure the results of this study, task completion rate was used, which has the value of 1 only if the participant solves the task correctly, 0 otherwise. Overall, the majority of the tasks were completed successfully by $\ge80\%$ of participants (see Figure~\ref{user_study_results}). The tasks T2 and T6 represent exceptions with the overall worst completion rate, solved correctly by 30\% and 40\% of participants respectively. 

After completing the tasks, the participants were asked to fill out the System Usability Scale (SUS) \cite{Brooke1996} questionnaire. It consists of ten statements that the study participants ranked on a 5-point Likert scale ranging from 1 (strongly disagree) to 5 (strongly agree). The ranking of the statements are then used to calculate the SUS score which expresses the usability of the system. The value of the score ranges between 0 and 100, with higher values expressing better usability. The proposed application reached the SUS score of 81.25.

\subsection{Analysis of common Cholec80 dataset splits}
In order to validate the proposed framework, we analyze the most common splits \cite{Funke2023} of the Cholec80 dataset \cite{Twinanda2017} using our visualization framework and report our observations. We downsample phase annotations of the Cholec80 dataset to 1~fps to obtain frames with both phase and instrument labels. For simplicity, this analysis does not cover cross-validation splits.

\subsubsection{40/-/40 split}
\begin{figure}[h]
\centering
\includegraphics[width=\textwidth]{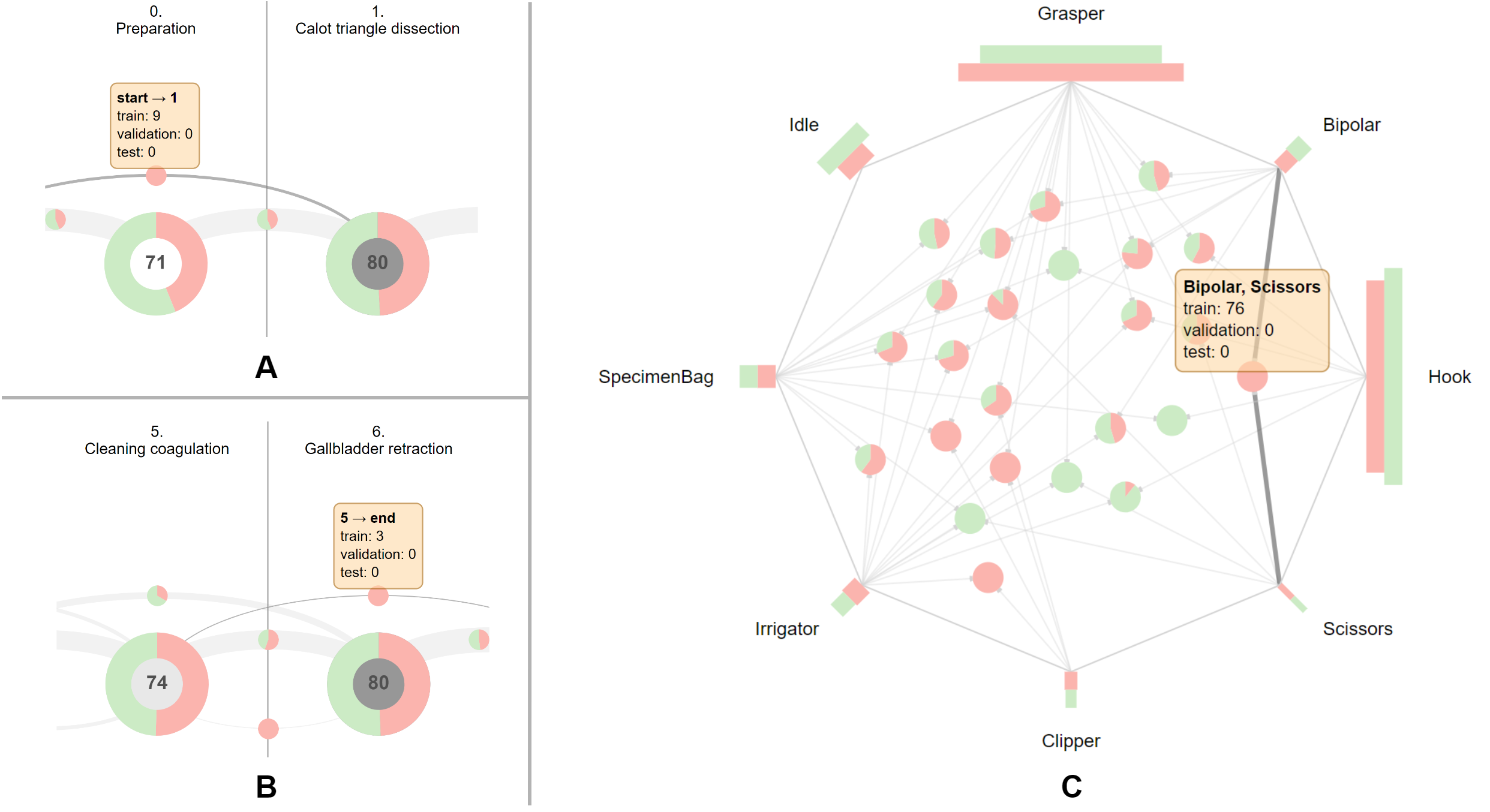}
\caption{Characteristics and shortcomings of the 40/-/40 split. Surgeries starting in the \textit{Calot triangle dissection} phase are only present in the training set (A). The ending sequence \textit{Gallbladder retraction} to \textit{Cleaning coagulation} occurs only in the training set (B). The instruments bipolar and scissors co-occur only in the training set (C).}\label{40_40_split}
\end{figure}

In the 40/-/40 split, which is used in the studies \cite{Twinanda2017,Chen2018}, all surgical phases are represented in both sets. However, a closer inspection of phase transitions unveils a group of nine surgeries (10, 13, 19, 22, 23, 29, 32, 33, 38) that deviate from the standard workflow by skipping the first phase and initiating the surgery directly in the second phase (see Figure~\ref{40_40_split}a). Notably, all of the nine surgeries are assigned to the training set, therefore the evaluation of the model's performance on the test set does not include this special workflow. In addition, another unique workflow that only occurs in three surgeries (12, 14, 32) in the training set can be identified using the proposed visualization (see Figure~\ref{40_40_split}b). After the \textit{Gallbladder packaging} phase, these three surgeries move on to the \textit{Gallbladder retraction}, thus omitting the \textit{Cleaning coagulation} phase. Subsequently, the surgeries return to the previously skipped \textit{Cleaning coagulation} phase which is also the final phase of the three surgeries. Since this unique sequence of phases only appears in the training set, they are not included in the evaluation of the machine learning model. With this information at hand, the split can be optimized by re-assigning the surgeries 32, 33, and 38 to the test set, as interactively determined in our tool. Accordingly, three randomly selected surgeries 58, 66, and 71 from the test set are assigned to the training set to retain the 40/-/40 split. As a result of this re-partition, the aforementioned cases of phase transitions now also appear in the test set.

Regarding the instrument use, the proposed visualization shows that all of the individual instruments are represented in all sets and also follow similar distributions. Nevertheless, there are several instrument combinations that do not occur in one of the sets (see Figure~\ref{40_40_split}c). However, these instruments combinations mostly represent rare cases, as they account for only a small fraction of the dataset and appear in single surgeries.

\subsubsection{32/8/40 split}
\begin{figure}[h]
\centering
\includegraphics[width=\textwidth]{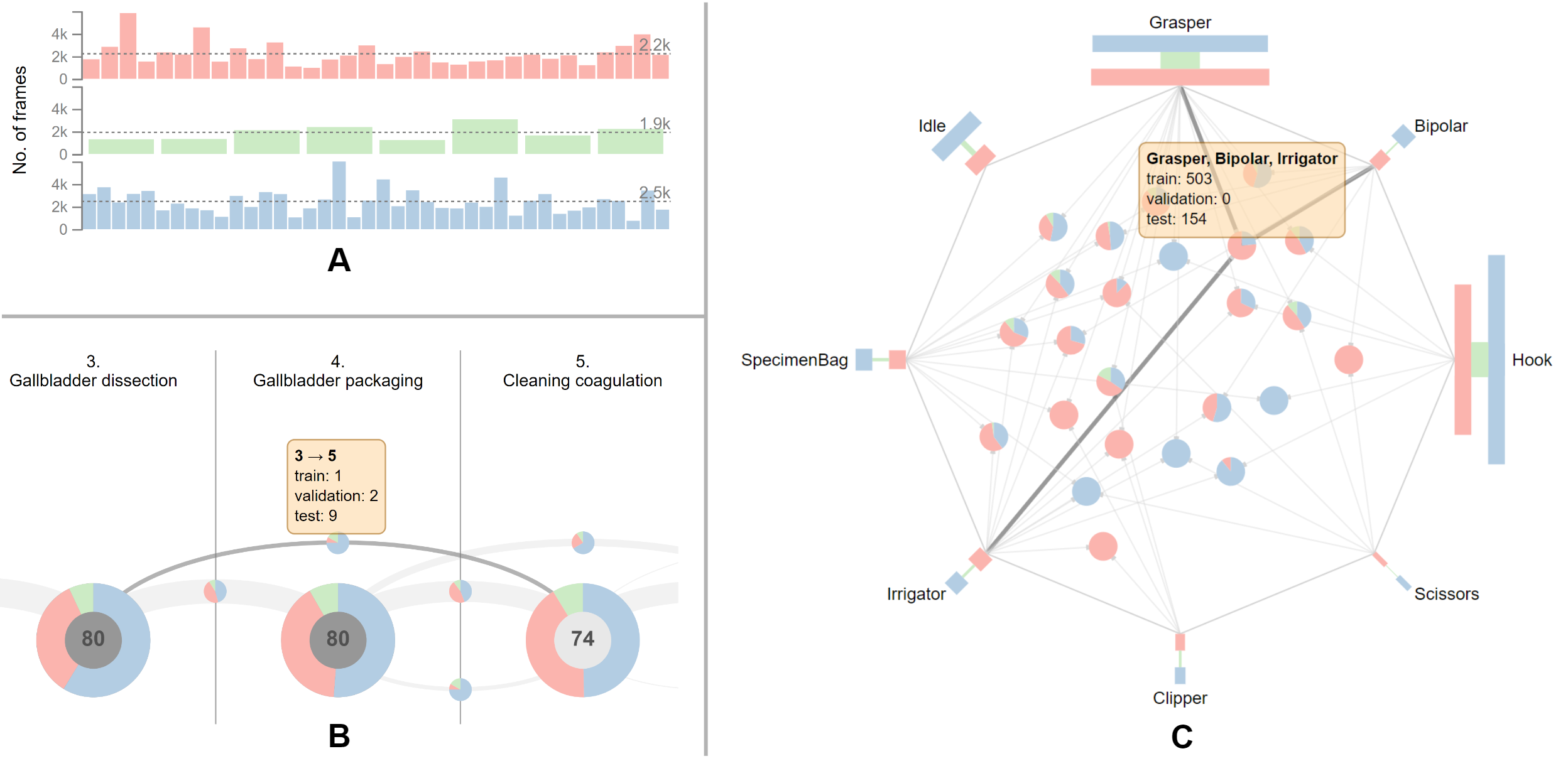}
\caption{Characteristics and shortcomings of the 32/8/40 split. Surgeries from the validation set have fewer frames on average, compared to the training and test sets (A). The phase transitions (\textit{Gallbladder dissection}, \textit{Cleaning coagulation}) and (\textit{Cleaning coagulation}, \textit{Gallbladder packaging}) occur only once in the training set (B). The simultaneous occurrence of the instruments grasper, bipolar, and irrigator is not represented in the validation set (C).}\label{32_8_40_split}
\end{figure}

To perform model selection or hyperparameter search, studies \cite{Jin2018,Jin2020,Czempiel2020,Zou2022} use eight surgeries from the training set for validation, resulting in a 32/8/40 split. This split yields sufficient representation of phases across sets. However, surgeries from the validation set have fewer frames on average ($\approx 1,900$ frames) than the training and test sets with $\approx 2,200$ and $\approx 2,500$ frames respectively (see Figure~\ref{32_8_40_split}a). Especially, the disparity between the average duration of surgeries from the validation and test set ($\approx 10$ min) might affect the performance estimation on these sets. As the surgery duration can indicate its complexity, the surgeries from the validation set may be easier to infer.

Similar to the 40/-/40 split, the surgeries skipping the first phase are found exclusively in the training and validation sets. This can be solved with our tool by assigning the surgeries 10 and 13 from the training to the test set, and randomly selected surgeries 48 and 64 from the test to the training set. Besides, the 32/8/40 split entails reduction of the training set size. This becomes especially apparent in case of two phase transitions (\textit{Gallbladder dissection}, \textit{Cleaning coagulation}) and (\textit{Cleaning coagulation}, \textit{Gallbladder packaging}) as they are reduced from three occurrences to just a single occurrence in the training set, as opposed to two and nine occurrences in the validation and test set respectively (see Figure~\ref{32_8_40_split}b). This will presumably hinder the generalization of the model. Regarding the instruments, the co-occurrences of surgical instruments that are missing in one of the sets are more prevalent in this split due to the additional validation set. One considerable example is the simultaneous use of grasper, bipolar, and irrigator occurring in 503 frames in the training set and in 154 frames in the test set (see Figure~\ref{32_8_40_split}c).

\subsubsection{40/8/32 split}
Instead of setting aside eight surgeries from the training set, two studies \cite{Czempiel2020, Rivoir2023} select eight surgeries from the testing set for validation, thus creating a 40/8/32 split. In this split, all phases as well as single instruments are present in all sets and also follow similar distributions. Similar to the original 40/-/40 split, surgeries starting in the \textit{Calot triangle dissection} phase, are exclusive to the training set. Furthermore, the three surgeries that move on from  \textit{Gallbladder packaging} to \textit{Gallbladder retraction} and end in the \textit{Cleaning coagulation} phase are also found only in the training set. This particular issue can be addressed by moving the surgery 32 to the validation set, the surgeries 33 and 38 to the test, and randomly selected surgeries 46, 58, and 70 to the training set to retain the 40/8/32 split.

\begin{figure}[h]
\centering
\includegraphics[width=\textwidth]{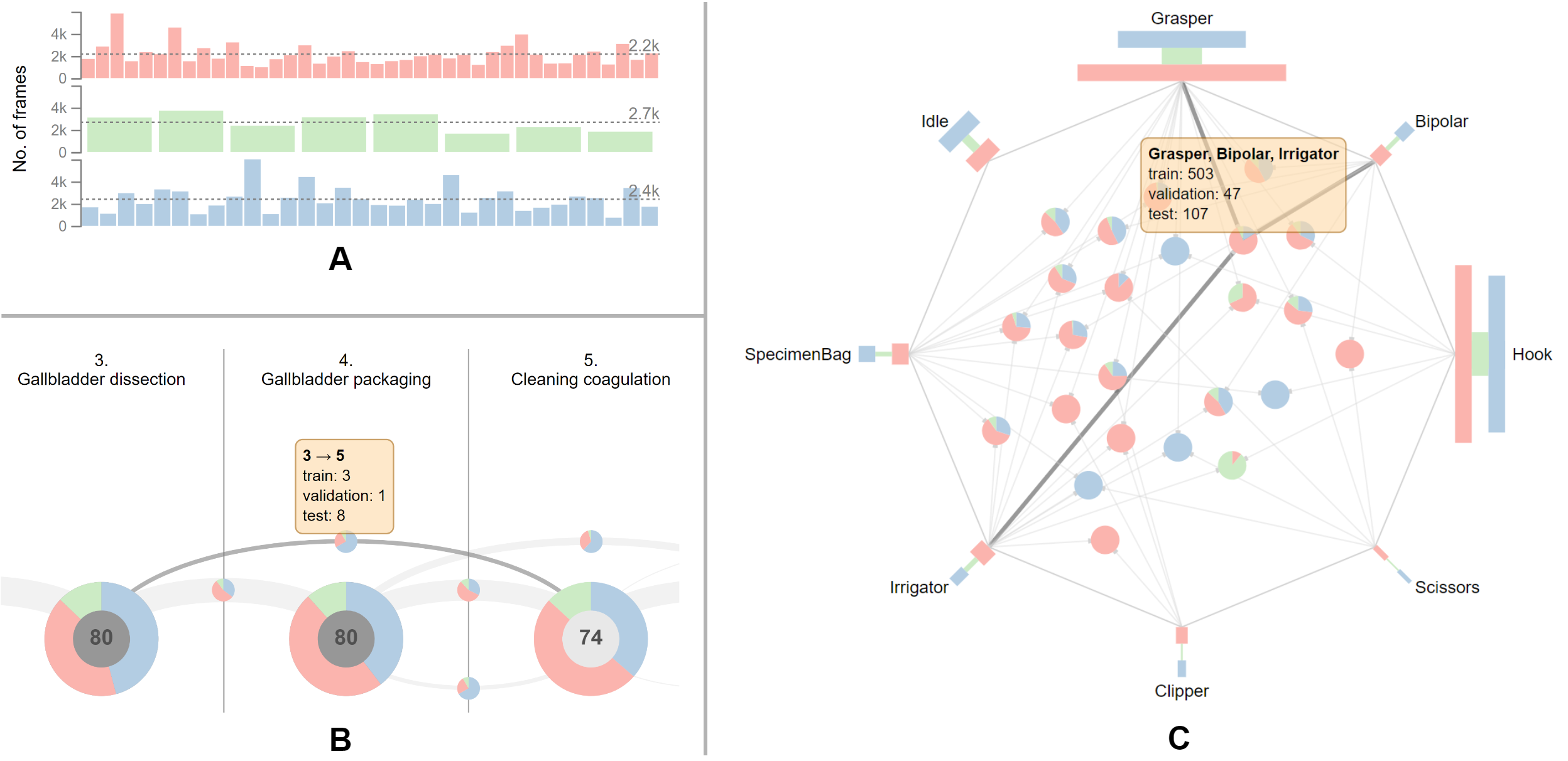}
\caption{Characteristics of the 40/8/32 split. Surgeries from the validation set contain more frames on average than surgeries from other sets (A). Furthermore, this split provides a better coverage of the phase transitions (\textit{Gallbladder dissection}, \textit{Cleaning coagulation}) and (\textit{Cleaning coagulation}, \textit{Gallbladder packaging}) in the training set, compared to the 32/8/40 split (B). The combination of grasper, bipolar, and irrigator appears in all sets (C).}\label{40_8_32_split}
\end{figure}

Compared to the 32/8/40 split, the validation set holds a larger amount of frames, thus resulting in a better coverage of various cases (see Figure~\ref{40_8_32_split}a). Furthermore, the phase transitions (\textit{Gallbladder dissection}, \textit{Cleaning coagulation}) and (\textit{Cleaning coagulation}, \textit{Gallbladder packaging}) now appear three times in the training set, thus providing more samples for the training of the model (see Figure~\ref{40_8_32_split}b). Considering the co-occurrence of instruments, an improvement over the 32/8/40 split can be observed, as the combination of grasper, bipolar, and irrigator now also appears on 47 frames in the validation set (see Figure~\ref{40_8_32_split}c).

\subsection{Summary of unrepresented cases}
Table~\ref{unrepresented_cases} shows common dataset splits of the Cholec80 dataset as well as the number of phase transitions, intra-phase instruments, and instrument combinations that are not represented in one of the sets. As can be seen in the table, training set is unaffected by different split variants, however, the number of unrepresented cases in validation and test sets dramatically changes on different splits. Visualizations of the splits are provided in the supplementary information. 

\begin{table}[h]
\begin{center}
\begin{minipage}{\textwidth}
\caption{Number of unrepresented cases in the training, validation, and test sets that were discovered using the proposed visualization framework.}\label{unrepresented_cases}
\begin{tabular*}{\textwidth}{@{\extracolsep{\fill}}ccccccccccc@{\extracolsep{\fill}}}

\toprule
 & & \multicolumn{9}{c}{Unrepresented cases} \\
\cmidrule{3-11}
 & & \multicolumn{3}{c}{\shortstack{Phase\\transition}} & \multicolumn{3}{c}{\shortstack{Instrument\\during phase}} & \multicolumn{3}{c}{\shortstack{Instrument\\combination}} \\
 
\cmidrule{3-5}\cmidrule{6-8}\cmidrule{9-11}

Split & Publications & \multirow{2}{*}{\rotatebox[origin=c]{90}{Train}} & \multirow{2}{*}{\rotatebox[origin=c]{90}{Val}} & \multirow{2}{*}{\rotatebox[origin=c]{90}{Test}} & \multirow{2}{*}{\rotatebox[origin=c]{90}{Train}} & \multirow{2}{*}{\rotatebox[origin=c]{90}{Val}} & \multirow{2}{*}{\rotatebox[origin=c]{90}{Test}} & \multirow{2}{*}{\rotatebox[origin=c]{90}{Train}} & \multirow{2}{*}{\rotatebox[origin=c]{90}{Val}} & \multirow{2}{*}{\rotatebox[origin=c]{90}{Test}} \\
\tiny{(train/val/test)} & & & & & & & & & & \\

\midrule
40/-/40 & \cite{Twinanda2017,Chen2018} & 0 & - & 3 & 4 & - & 3 & 4 & - & 4 \\
32/8/40 & \cite{Jin2018,Jin2020,Czempiel2020,Zou2022} & 0 & 2 & 3 & 4 & 17 & 3 & 4 & 14 & 4 \\
40/8/32 & \cite{Czempiel2020,Rivoir2023} & 0 & 3 & 3 & 4 & 11 & 6 & 4 & 9 & 6 \\
40/24/16 & \cite{Liu2022} & 0 & 3 & 3 & 4 & 4 & 14 & 4 & 6 & 9 \\
\botrule
\end{tabular*}
\end{minipage}
\end{center}
\end{table}

\section{Discussion and future work}
This work presents a publicly available visualization framework that facilitates interactive assessment of dataset splits for surgical phase and instrument recognition. The motivation for this has been previously outlined in some studies. \mbox{Zisimopoulos et al.} \cite{Zisimopoulos2018} report a high discrepancy of the model's performance on validation and test sets which is attributed to some phases missing in the validation set. Moreover, the effects of imbalances of instrument co-occurrences on model's performance have been previously highlighted by \mbox{Sahu et al.} \cite{Sahu2017}. The visualization framework presented in this work is specifically designed to address these cases.

Using our tool, we were able to pinpoint several aspects of the splits that can distort the evaluation of the model's performance. Moreover, the application enabled us to eliminate some of these issues by manually re-partitioning the sets. Furthermore, we discovered that the number of unrepresented cases significantly varies for different dataset splits. In future work, algorithms for the generation of optimal dataset splits \cite{Vakayil2022} can be explored. The user study shows promising results, as most of the tasks were correctly solved by at least 80\% of the participants. The positive outcome of the user study is further supported by the SUS score of 81.25 which indicates above average usability of the system.

The scope of this application is limited to the analysis of phase and instrument annotations. However, visual features, such as bad lighting conditions, over or underexposed instruments, and occlusions have high influence on the performance of the model \cite{Jin2018}. Future work should also include analysis of distributions of various visual features. Correspondingly, it can be also extended to support adjacent tasks including instrument and pathology detection or segmentation with bounding-box or pixel-level predictions. Finally, we also believe that integration of more fine-grained surgical activity information, such as action triplets \cite{Sharma2023}, can provide a more sophisticated overview of surgical workflows.

\section{Conclusion}\label{conclusion}
In this work, we presented a publicly available application implemented for the research community that aims to facilitate visual exploration of dataset splits for surgical phase and instrument recognition. To validate the design of our application, we conducted a user study with ten participants. Further, we performed an analysis of the most common Cholec80 splits and identified improvements of the splits using our tool. The results indicate that the proposed application can enhance the development process of machine learning models for surgical phase recognition by providing insights into the dataset splits, potentially resulting in more reliable performance evaluations. Furthermore, we believe that organizers of biomedical challenges can also greatly benefit from the proposed framework during the preparation of challenge datasets.


\backmatter

\bmhead{Supplementary information}
The supplementary information is provided. Source code is available at \url{https://github.com/Cardio-AI/endovis-ml} and the live application can be accessed at \url{https://cardio-ai.github.io/endovis-ml/}. 

\bmhead{Acknowledgments} We thank the participants of the user study for their contribution to this work. 

\section*{Declarations}

\bmhead{Conflict of interest}
The authors declare that they have no conflict of interest.
\bmhead{Funding}
This work was supported by Informatics for Life funded by the Klaus Tschira Foundation.
\bmhead{Ethical approval}
All procedures performed in studies involving human participants were in accordance with the ethical standards of the institutional and/or national research committee and with the 1964 Helsinki declaration and its later amendments or comparable ethical standards.
\bmhead{Informed consent}
Informed consent was obtained from all individual participants included in the study.

\bibliography{bibliography}

\clearpage
\setcounter{section}{0}
\setcounter{figure}{0}
\setcounter{table}{0}

\section*{Supplementary information}
\section{User study}
The following Table~\ref{user_study_tasks} shows tasks from the user study of the presented visualization framework.

\begin{table}[h]
\begin{center}
\caption{List of tasks to be solved by study participants using the  visualization framework.}\label{user_study_tasks}
\begin{tabular}{@{}ll@{}}
\toprule
ID  & Description                                                       \\
\midrule
T1  & Which phase has the most frames?                                  \\
T2  & Which two phases have the largest proportion of idle frames?      \\
T3  & Which instruments co-occur with the Clipper?                      \\
T4  & Which phase is not represented in all dataset splits?             \\
T5  & Which phases are not present in all surgeries?                    \\
T6  & Which phase transitions are not present in the training set?      \\
T7  & In which phase are the Scissors used most often?                  \\
T8  & Which surgeries end in the Cleaning coagulation phase?            \\
T9  & Which instrument combination is not present in the training set?  \\
T10 & In which phase do Bipolar, Irrigator and SpecimenBag co-occur?    \\
\botrule
\end{tabular}
\end{center}
\end{table}

Three statements from the SUS questionnaire that received the most positive rating.

\begin{itemize}
    \item Q4: I think that I would need the support of a technical person to be able to use this system (mean 1.2)
    \item Q5: I found the various functions in this system were well integrated (mean 4.5)
    \item Q6: I thought there was too much inconsistency in this system (mean 1.4)
\end{itemize}

Conversely, we report three statements that were rated least favorably.

\begin{itemize}
    \item Q3: I thought the system was easy to use (mean 3.9) 
    \item Q7: I would imagine that most people would learn to use this system very quickly (mean 3.9)
    \item Q10: I needed to learn a lot of things before I could get going with this system (mean 2.1)
\end{itemize}

\pagebreak

\section{Analysis of common Cholec80 dataset splits}
\begin{sidewaysfigure}
\includegraphics[width=\textwidth]{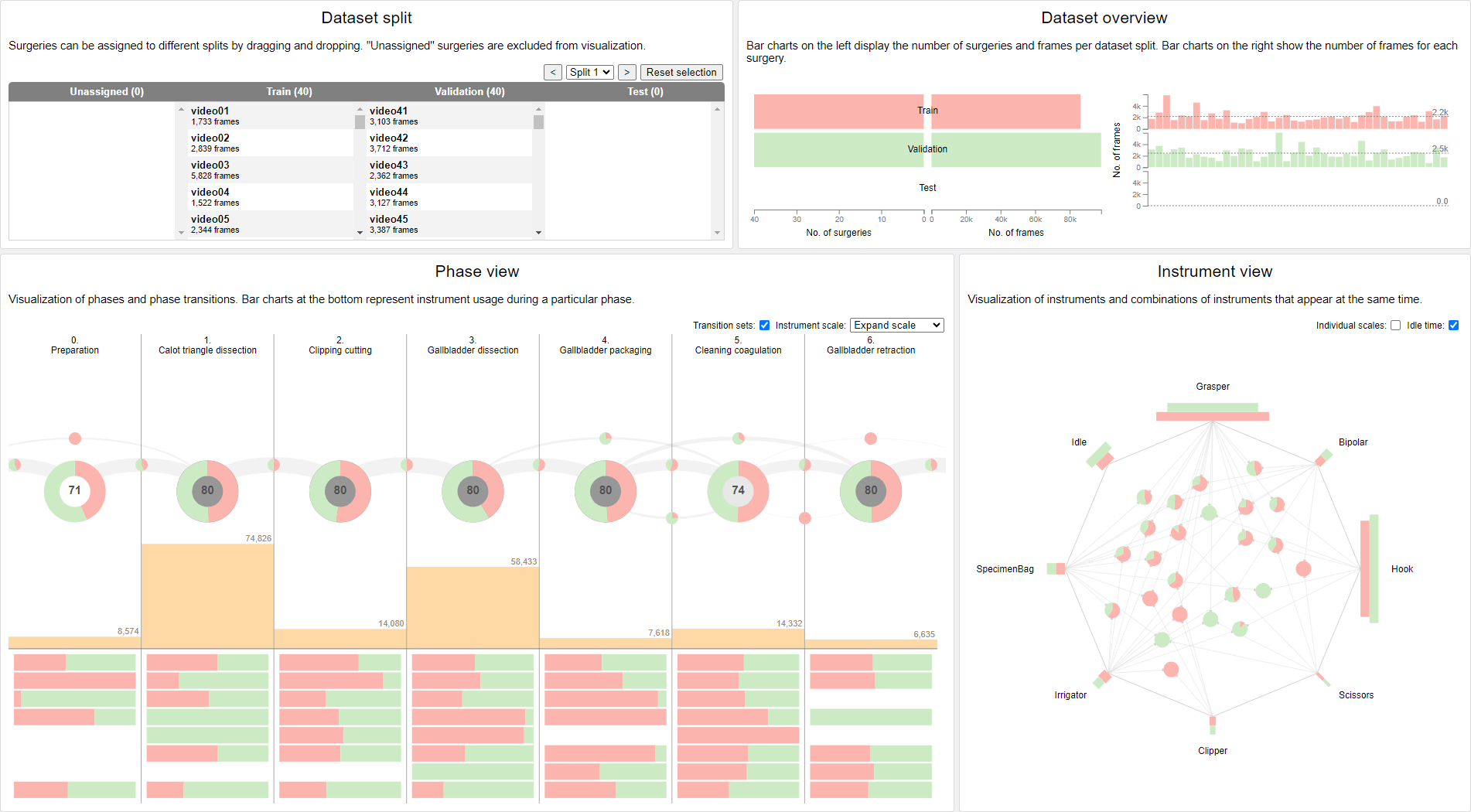}
\caption{Screenshot of the application with the 40/-/40 split  of the Cholec80 dataset.}\label{40_40}
\end{sidewaysfigure}

\begin{sidewaysfigure}
\includegraphics[width=\textwidth]{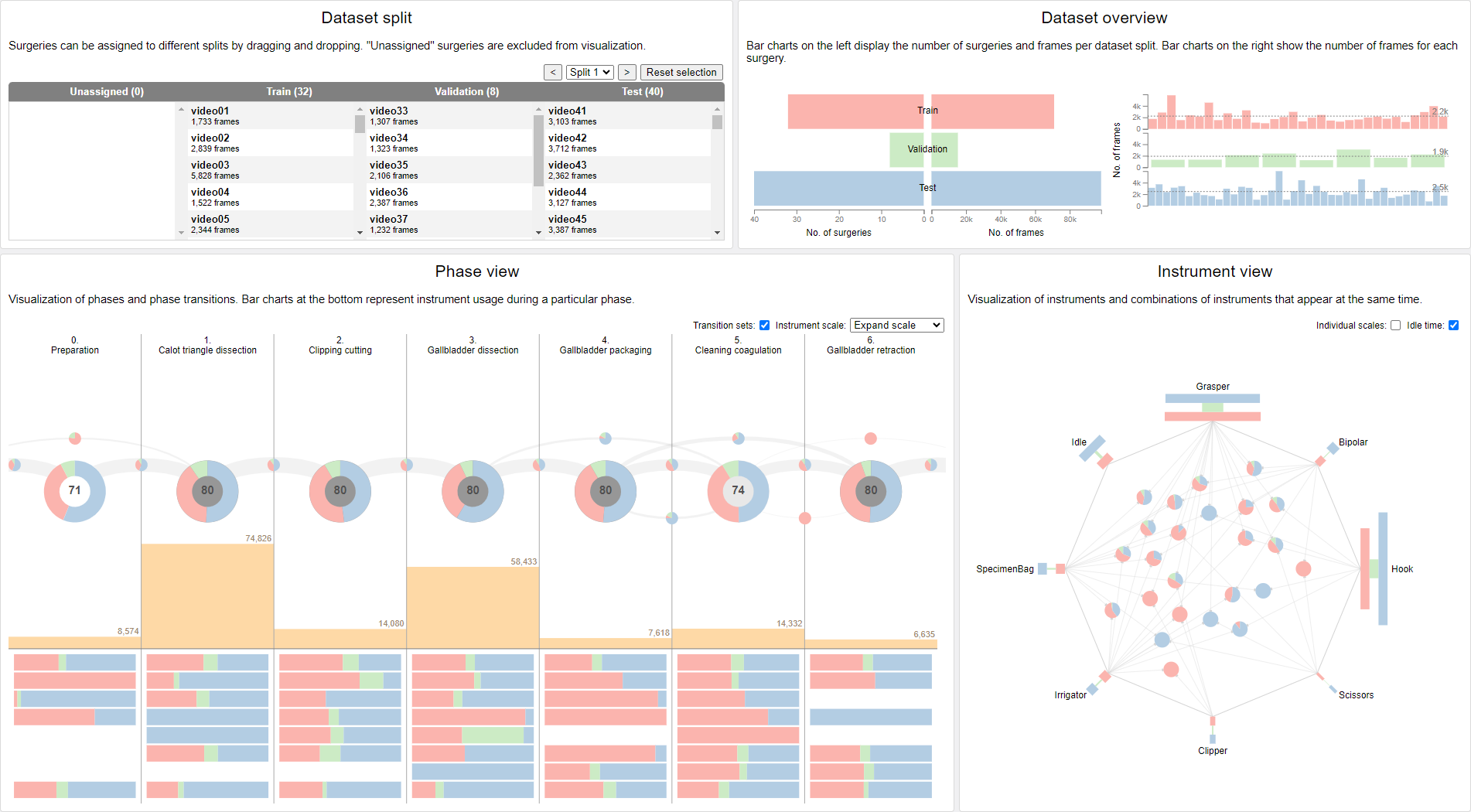}
\caption{Screenshot of the application with the 32/8/40 split  of the Cholec80 dataset.}\label{32_8_40}
\end{sidewaysfigure}

\begin{sidewaysfigure}
\includegraphics[width=\textwidth]{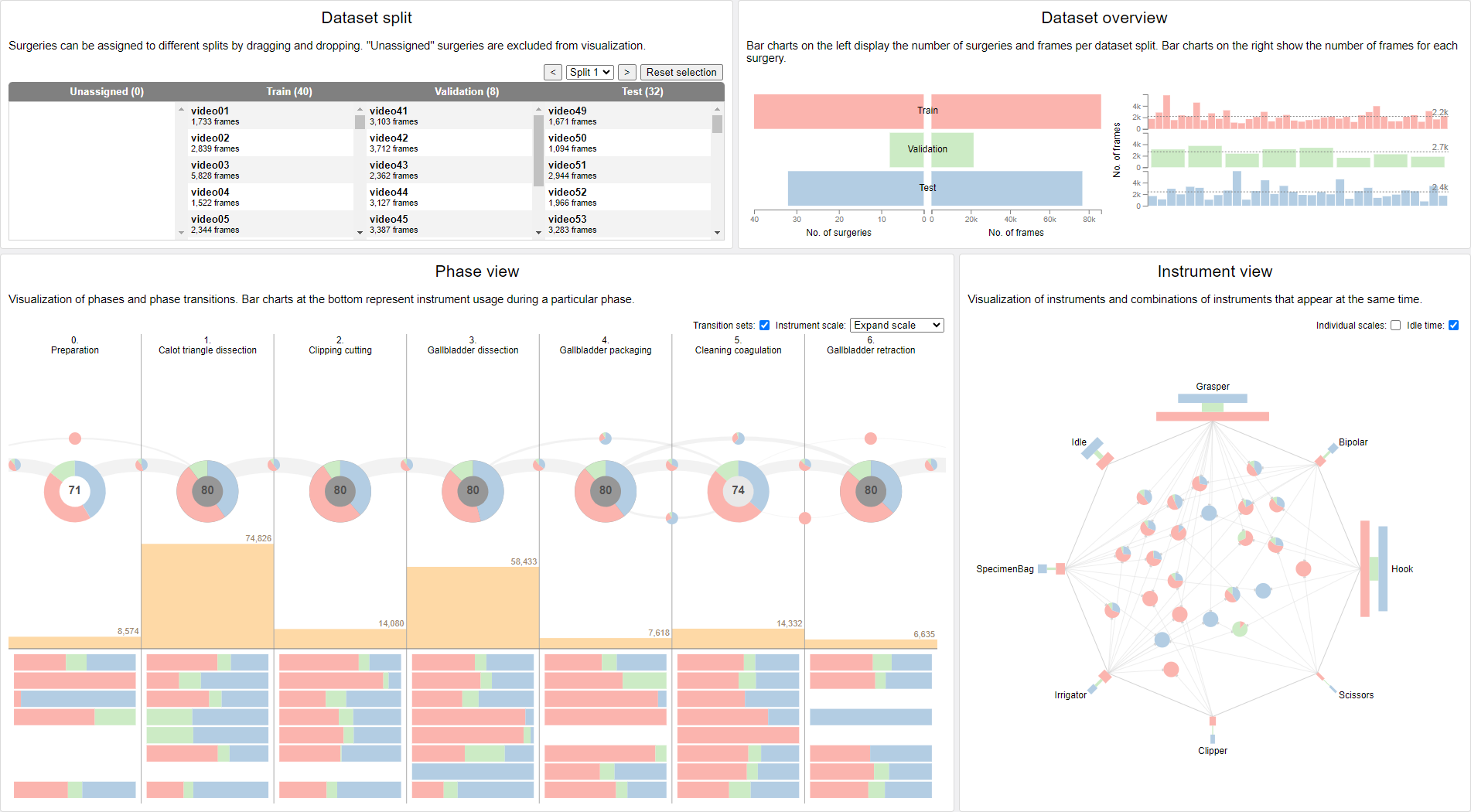}
\caption{Screenshot of the application with the 40/8/32 split  of the Cholec80 dataset.}\label{40_8_32}
\end{sidewaysfigure}

\begin{sidewaysfigure}
\includegraphics[width=\textwidth]{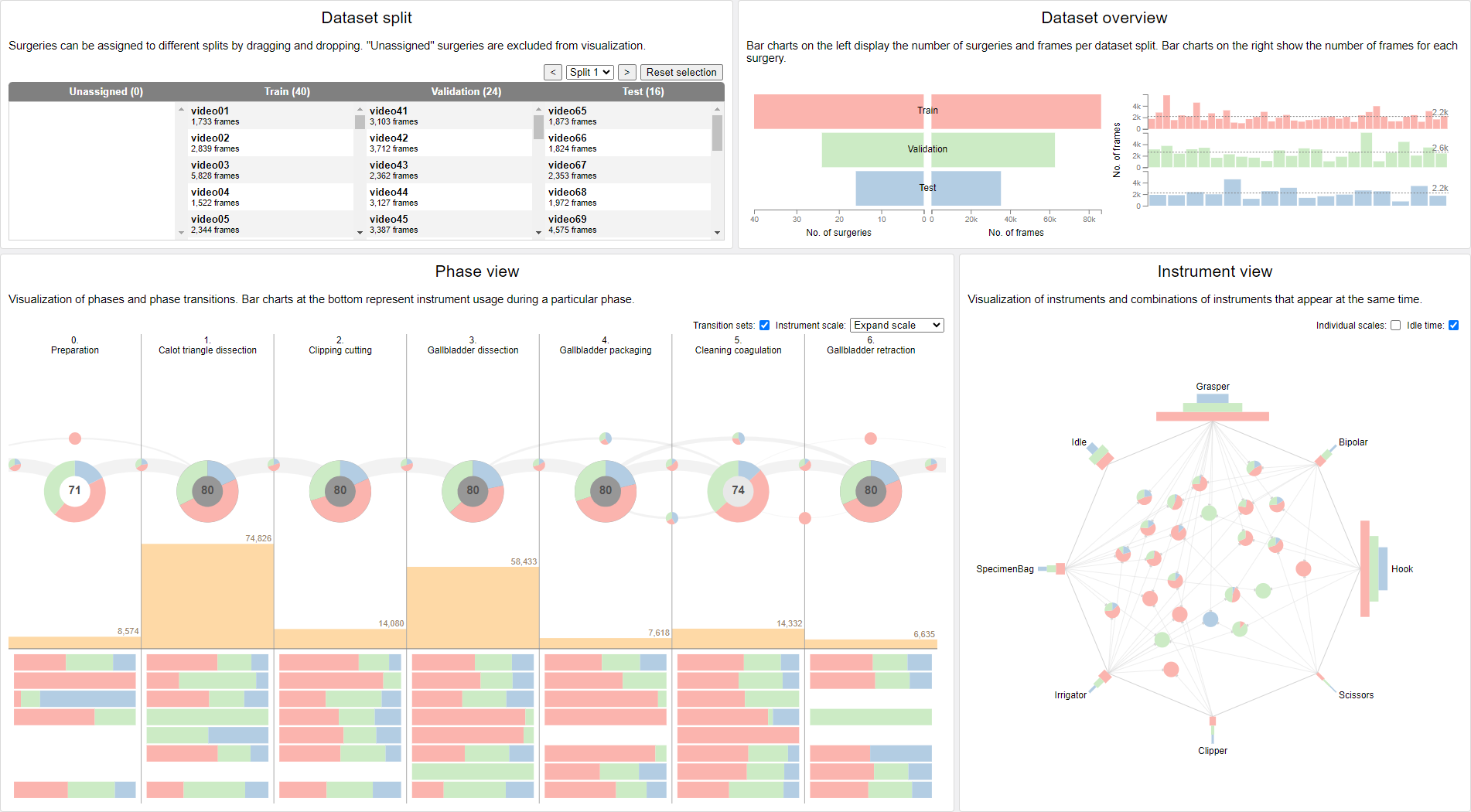}
\caption{Screenshot of the application with the 40/24/16 split of the Cholec80 dataset.}\label{40_24_16}
\end{sidewaysfigure}

The following Table~\ref{split_ids} shows common Cholec80 dataset splits as well as the allocation of surgeries to sets. Further, we provide visualization of these splits in Figure~\ref{40_40}, Figure~\ref{32_8_40}, Figure~\ref{40_8_32}, and Figure~\ref{40_24_16}.

\begin{table}[h]
\begin{center}
\begin{minipage}{\textwidth}
\caption{Most common Cholec80 dataset splits and the allocation of surgeries to sets.}\label{split_ids}
\begin{tabular*}{\textwidth}{@{\extracolsep{\fill}}llp{8cm}@{\extracolsep{\fill}}}
\toprule
Split & Set & File IDs \\
\midrule
\multirow{7}{*}{40/40}      & \multirow{3}{*}{Training}   & 1, 2, 3, 4, 5, 6, 7, 8, 9, 10, 11, 12, 13, 14, 15, 16, 17, 18, 19, 20, 21, 22, 23, 24, 25, 26, 27, 28, 29, 30, 31, 32, 33, 34, 35, 36, 37, 38, 39, 40 \\
\cmidrule{2-3}
                            & \multirow{3}{*}{Test}       & 41, 42, 43, 44, 45, 46, 47, 48, 49, 50, 51, 52, 53, 54, 55, 56, 57, 58, 59, 60, 61, 62, 63, 64, 65, 66, 67, 68, 69, 70, 71, 72, 73, 74, 75, 76, 77, 78, 79, 80 \\

\midrule

\multirow{7}{*}{32/8/40}    &\multirow{2}{*}{Training}      & 1, 2, 3, 4, 5, 6, 7, 8, 9, 10 ,11, 12, 13, 14, 15, 16, 17, 18, 19, 20, 21, 22, 23, 24, 25, 26, 27, 28, 29, 30, 31, 32 \\
\cmidrule{2-3}
                            &\multirow{1}{*}{Validation}    & 33, 34, 35, 36, 37, 38, 39, 40 \\
\cmidrule{2-3}
                            &\multirow{2}{*}{Test}          & 41, 42, 43, 44, 45, 46, 47, 48, 49, 50, 51, 52, 53, 54, 55, 56, 57, 58, 59, 60, 61, 62, 63, 64, 65 , 66, 67, 68, 69, 70, 71, 72, 73, 74, 75, 76, 77, 78, 79, 80 \\

\midrule
\multirow{7}{*}{40/8/32}    & \multirow{3}{*}{Training}    & 1, 2, 3, 4, 5, 6, 7, 8, 9, 10, 11, 12, 13, 14, 15, 16,17, 18, 19, 20, 21, 22, 23, 24, 25, 26, 27, 28, 29, 30, 31, 32, 33, 34, 35, 36, 37, 38, 39, 40 \\
\cmidrule{2-3}
                            & \multirow{1}{*}{Validation}  & 41, 42, 43, 44, 45, 46, 47, 48 \\
\cmidrule{2-3}
                            & \multirow{2}{*}{Test}        & 49, 50, 51, 52, 53, 54, 55, 56, 57, 58, 59, 60, 61, 62, 63, 64 , 65, 66, 67, 68, 69, 70, 71, 72, 73, 74, 75, 76, 77, 78, 79, 80 \\

\midrule

\multirow{7}{*}{40/24/16}  & \multirow{3}{*}{Training}     & 1, 2, 3, 4, 5, 6, 7, 8, 9, 10, 11, 12, 13, 14, 15, 16, 17, 18, 19, 20, 21, 22, 23, 24, 25, 26, 27, 28, 29, 30, 31, 32, 33, 34, 35, 36, 37, 38, 39, 40 \\
\cmidrule{2-3}
                            & \multirow{2}{*}{Validation}   & 41, 42, 43, 44, 45, 46, 47, 48, 49, 50, 51, 52, 53, 54, 55, 56, 57, 58, 59, 60, 61, 62, 63, 64 \\
\cmidrule{2-3}
                            & \multirow{1}{*}{Test}         & 65, 66, 67, 68, 69, 70, 71, 72, 73, 74, 75, 76, 77, 78, 79, 80 \\
\botrule
\end{tabular*}
\end{minipage}
\end{center}
\end{table}

\end{document}